\documentclass[sigconf]{acmart}

\usepackage[utf8]{inputenc} 
\usepackage[T1]{fontenc}    
\usepackage{hyperref}       
\usepackage{url}            
\usepackage{booktabs}       
\usepackage{amsfonts}       
\usepackage{nicefrac}       
\usepackage{microtype}      
\usepackage{xcolor}         
\usepackage{graphicx}
\usepackage{amsmath}
\usepackage{multirow}
\usepackage{caption}
\usepackage{subcaption}
\usepackage{ulem}
\usepackage{booktabs}
\usepackage{multirow}

\newcommand{\sz}[1]{\textcolor{black}{#1}}

\AtBeginDocument{%
  }

\setcopyright{acmcopyright}

\acmConference[MLG'22]{International Workshop on Mining and Learning with Graphs}{August 15,
  2022}{Washington DC}

\begin{document}

\title{Learning Personalized Representations using Graph Convolutional Network}

\author{Hongyu Shen}
\authornote{Work is done during the internship at Amazon Alexa AI.}
\affiliation{%
  \institution{Amazon Alexa}
  \city{Seattle}
  \state{WA}
  \country{USA}
}
\email{hongyus@amazon.com}

\author{Jinoh Oh}
\affiliation{%
    \institution{Amazon Alexa}
  \city{Seattle}
  \state{WA}
  \country{USA}}
\email{ojino@amazon.com}

\author{Shuai Zhao}
\affiliation{%
    \institution{Amazon Alexa}
  \city{Seattle}
  \state{WA}
  \country{USA}
}
\email{shuzhao@amazon.com}

\author{Guoyin Wang}
\affiliation{%
   \institution{Amazon Alexa}
  \city{Seattle}
  \state{WA}
  \country{USA}}
\email{guoyiwan@amazon.com}

\author{Tara Taghavi}
\affiliation{%
    \institution{Amazon Alexa}
  \city{Seattle}
  \state{WA}
  \country{USA}}
\email{taghavit@amazon.com}

\author{Sungjin Lee}
\affiliation{%
    \institution{Amazon Alexa}
  \city{Seattle}
  \state{WA}
  \country{USA}}
\email{sungjinl@amazon.com}



\renewcommand{\shortauthors}{Trovato et al.}


\begin{abstract}
\label{abs}
Generating representations that precisely reflect customers’ behavior is an important task for providing personalized skill routing experience in Alexa. Currently, Dynamic Routing (DR) team, \sz{which} is responsible for routing Alexa traffic to providers/skills, relies on two features to be served as personal signals: absolute traffic count and normalized traffic count of every skill usage per customer. Neither of them considers the network-structure for interactions between customers and skills, which contain richer information for customer preferences. In this work, we first build a heterogeneous edge-attributed graph based customers’ past interactions with the invoked skills, in which the user requests (utterances) are modeled as edges. Then we propose a graph convolutional network(GCN)-based model, namely Personalized Dynamic Routing Feature Encoder (PDRFE), that generates personalized customer representations learned from the built graph. Compared with existing models, PDRFE is able to further capture contextual information in the graph convolutional function. The performance of our proposed model is evaluated by a downstream task, defect prediction, that predicts the defect label from the learned embeddings of customers and their triggered skills. We observe up to 41\% improvements on the cross-entropy metric for our proposed models compared to the baselines.

\end{abstract}





\maketitle

\section{Introduction}
\label{intro}
The advancement of Alexa AI technologies \sz{enables} customers to have various options for interacting with Alexa devices. For example, a person can ask Echo to play a song, to turn on a TV, or to open air conditioners while being away from home. Alexa provides such convenience to help people automate daily tasks with ease. So far, Alexa AI has created more than 200k different skills that range from video display, automatic subscriptions to home services and etc. ~\sz{Skills are similar to apps to process each customer query request.} Such services cover almost every possible task in customer's daily life. Providing such high-end services with Alexa has also attracted nearly 800 million registered users generating about 500 million user-Alexa interactions daily.

One of the core concepts behind Alexa services is the representation learning models created by Alexa Dynamic Routing (Alexa DR) team. Alexa DR develops ranking models and rules to generate representations of the skills and use these representations to rank each skill for every customer's request. Subsequently, customers' requests can be routed to the skills (e.g. the internal function in Alexa that provides video display or weather broadcasting) that can response to those requests.
Recently, Alexa DR starts to extend its ranking models to consider customer-skill interactions by using simple features such as normalized impression counts of successful customer interactions with skills as input features to its ranking models. However, these simple features are not strong enough to fully encode the complex nature of customer-skill interactions. For example, a simple impression count hardly captures the similarity between the skills, and thus looses its predictive power in that regard.
Another pitfall is that simple features are computed and used without any consideration on the context. For example, the normalized traffic count for a pair of customer and skill is invariant to the utterance text. 

In fact, previous studies have widely demonstrated the superior performance of applying graph modeling to describe interactions between two or more entities~\cite{DBLP:journals/debu/HamiltonYL17,schlichtkrull2018modeling,wang2018deep,velivckovic2017graph,wang2019kgat,wang2019heterogeneous}. This is similar to modeling customer-sill interactions in the form of a graph. there are also studies pointing out ~\sz{that} a graph encoded with ~\sz{contextual information} can further improve model performance from various perspectives~\cite{li2019spam,gong2019exploiting,hu2020heterogeneous,dwivedi2020generalization}. ~\sz{It} provides a natural mechanism to consider context information during modeling through edge-based convolution operations~\sz{, which} validates the attempt in combining the utterance text in modeling customer-skill relations.

However, there is one limitation ~\sz{when using the existing GCN models in our particular setup: they does not consider }the edge attribute in characterizing the directional between the customer and the skill nodes.
For example, with existing GCN-based models, the edge between customer and skill is predicted only by the customer and skill embedding or by a shared edge attribute that addresses equal weights on its connecting nodes. This means a customer will have the same edge probability to a certain skill regardless of the utterance text, which does not capture the true interaction for the probability of matching the customers to a certain skill should be different given different customers and the same utterance in general. Furthermore, the probabilities for the match of skills to a certain customer given different utterances should be different. The two existing issues in our particular setup suggests the need of a new GCN algorithm.

In this paper, we propose a model, named Personalized Dynamic Routing Feature Encoder (PDRFE), that aims to generate personalized embeddings for the customer nodes given the customer and his/her triggering utterances for different skills. PDRFE includes ~\sz{a \textit{personalizer} mechanism}, to provide customer embeddings adjusted with respect to the utterance text. Besides, we also consider two other modules in PDRFE to further improve the model performance by considering the edge features. The first one is the edge-based attention module. PDRFE applies attention module to distinguish the important connection and update the customer embedding and skill embedding accordingly. This is because some skills are designed to be more interactive than others, and create more frequent but less important request such as ``next'', ``repeat the question''. Thus, importance of edge needs to be distinguished and weighted differently to precisely capture the customer's true preference to skills. The second module is the edge convolution that considers all the edge features in updating the corresponding node embedding. In our case in Alexa, the edge features (e.g. utterances) indicate the customers behaviours to their invoked skills. And it also helps identifying the difference between skills. As a result, it is natural to consider the edge information in updating the node ~\sz{embeddings} for both the customers and skills. Overall, we compare our proposed model with 4 baselines ~\sz{through} a downstream evaluation task (i.e., defect prediction) that predicts the defect label from the personalized embedding of customers and the embedding of their triggering skills.
Additionally, we also provide ablation study for the modules we proposed in PDRFE to help identify the contribution of sub-modules to the downstream task.

\sz{The rest of paper is organized} as follows: We first discuss problem setup and the proposed PDRFE model in the next section, followed by the experiment section that includes experiment setup, the comparison results between PDRFE and the 4 baseline models through a downstream task evaluation, as well as ablation studies for PDRFE. Finally, we conclude our study and explicitly state the customer impact of this study for Amazon.

\begin{figure*}[ht]
\includegraphics[width=1.0\linewidth]{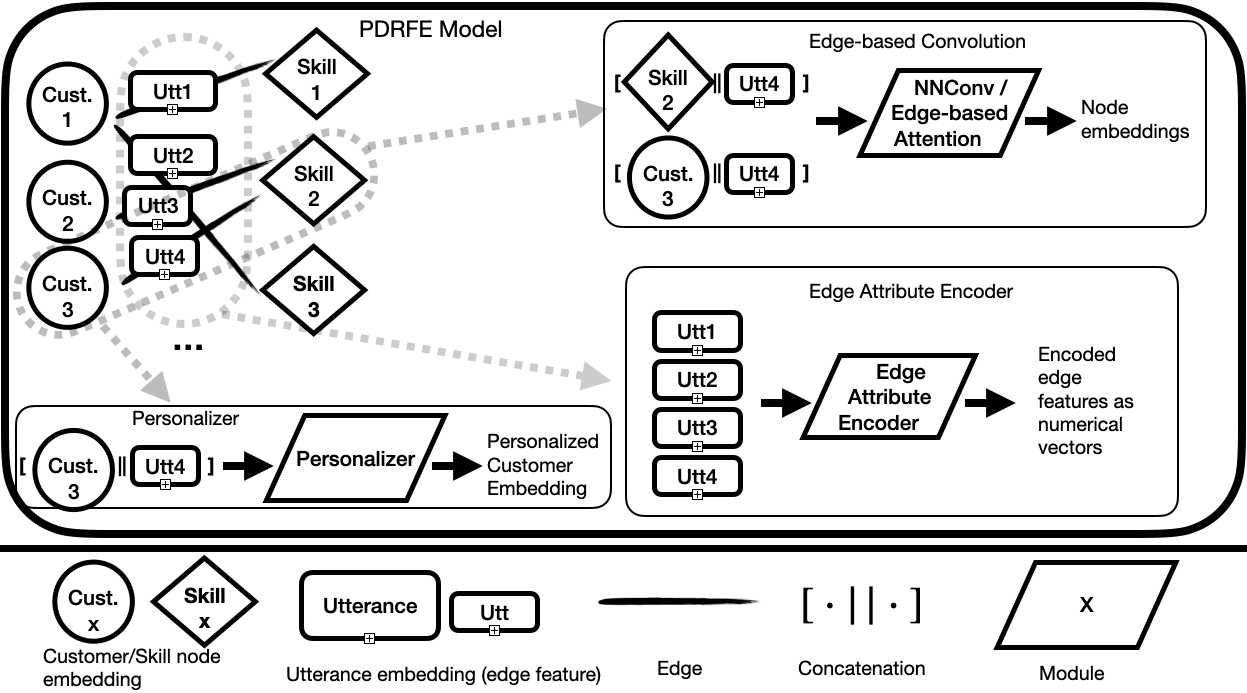}
\caption{Diagram of PDRFE Pipeline. There are three main components in the PDRFE pipeline. 1. The edge attribute encoder encodes all utterances to numerical vectors, which are the edge features in our graph. 2. The personalizer personalizes the customer node embedding by feeding the concatenation of customer node embedding and the utterance embedding to a personalizer module. 3. The Edge-based convoulution aggregates information concerning both the skill node embedding, personalized embedding and the utterance embedding.}
\label{fig:model_diagram}
\end{figure*}

\section{Problem Definition}
\label{problem_define}
Here we formally define our problem in the context of graph representation learning: 

Given a bipartite customer-skill interaction graph $G=(V,E)$ where $V$ is the set of nodes and $E$ is the set of edges and $V=\{U,S\}$ where $U$ refers to the set of customer nodes and $S$ refers to the set of skill nodes and $e_{u,s}$ denotes that a service request from user $u$ is fulfilled by skill $s$ ($e_{u,s} \in E$) (Note that, here we simplify the notation, but there can be multiple interactions between a customer $u$ and a skill $s$). For simplicity, we use $\textbf{e}_{u,s}$ to represent the feature vector to the edge $e_{u,s}$. The objective is to design a model $f$ that maps the initial input embeddings $(\textbf{h}_u^0, \textbf{e}_{u,s}, \textbf{h}_s^0)$ of a graph to a set of final embedding vectors for both the user and skill nodes, denoted as $(\textbf{h}_u^L, \textbf{h}_s^L) = f(\textbf{h}_u^0, \textbf{e}_{u,s}, \textbf{h}_s^0)$, such that when incorporating these embeddings to the downstream task, the corresponding objective function is optimized. Here the superscript $L$ for both $\textbf{h}_u$ and $\textbf{h}_s$ indicates the output embeddings of both customers and skills from the GCN model $f$.  We choose the objective for representation learning as link prediction.
In short, we have: 
\begin{description}
  \item[$\bullet$ Input:] $G$
  \item[$\bullet$ Output:] $\textbf{h}_u^L , \textbf{h}_s^L$ for all $u,s \in V$
  \item[$\bullet$ Link prediction:] \hfill \\ $\sum_{u\in U}\sum_{s \in \mathcal{N}^+(u), \hat{s} \in \mathcal{N}^-(u)} \max( M - <\textbf{h}_u^L, \textbf{h}_s^L> + <\textbf{h}_u^L, \textbf{h}_{\hat{s}}^L>, 0)\label{obj:link_prediction}$
  \item[$\bullet$ Downstream task (defect prediction):] \hfill \\ $\min_{\sigma} \frac{1}{|E|}\sum_{u \in U, s\ in S} CE(\sigma(\textbf{h}_u^L, \textbf{h}_s^L),y)$,
\end{description}
where $\sigma(\cdot,\cdot)$ is the defect classifier, and $<\cdot, \cdot>$ is the inner product. $\mathcal{N}^+(u)$ is positive neighbors ~\sz{---} the set of skill node that are connected to the customer $u$ node, and $\mathcal{N}^-(u)$ is negative neighbors - the set of randomly sampled noise skill nodes that are not connected to the customer node $u$ in the graph $G$. $M$ is a hyperparameter for the link prediction objective,  $y$ is defect label, and $CE(\cdot, \cdot)$ is the cross-entropy loss function.

\begin{table*}[ht]
  \caption{Fake Examples for Customer-Skill Interaction Log}
  \label{fake_data_tab}
  \centering
 \resizebox{\textwidth}{!}{
\begin{tabular}{lllll}
\toprule
Customer ID (CID) & Skill ID (SID) & Utterance (interaction1)  & Utterance (interaction2) & … \\ \midrule
10xxxxx           & Help           & What happened to my music & Can you help me          & … \\
21xxxxxx          & Help           & What can I do today       & Can you do a tutorial    & … \\
21xxxxxx          & Communication  & No                        & Connect to my phone      & … \\
89xxxxx           & Help           & Can you do a tutorial     & I have feedback          & … \\
10xxxxx           & Communication  & Drop in all devices       & Connect to my phone      & … \\
…                 & …              & …                         & …                        & … \\
    \bottomrule
  \end{tabular}}
\end{table*}

\section{Proposed Method}
\label{method}
Here we describe the structures of the proposed PDRFE model. Specifically, it involves 3 major components: 1) An encoder for edge attributes 2) An edge-based convolution operator in GCN that updates the embedding $(\textbf{h}_u, \textbf{h}_s)$ 3) A personalizer designed for generating personalized customer embedding that reflects ~\sz{a customer's request}. ~\sz{Overall, Figure~\ref{fig:model_diagram} illustrates the pipeline for the PDRFE model.} 

\subsection{Encoder for Edge Attributes}
In this work, the primary edge attribute is utterance text. We use a pre-trained Bi-LSTM text encoder to convert this utterance to feature vector, and there are three reasons.
First, our focus is about the interaction and not the semantic meaning of the utterance.
Thus, if it is carelessly trained from the scratch, there is a risk that the learned text encoder does not fully reflect semantic meaning of the utterance.
Second, using a pre-trained model reduces complexity in training and enables quicker iterations.
Third, using the Bi-LSTM over the other structures is to capture the cross-hypothesis correlation, which is an intermediate for generating utterance embeddings~\cite{kim2018scalable}.
Please note that this pre-trained model is ~\sz{widely used in Alexa}, hence the performance of this model has already been fully tested in production.

\subsection{Edge-based Convolution Operator}
Edge-based convolution operators allow the propagation of the edge features to all of its connected neighboring nodes. It is known that considering edge features during graph convolution usually enriches the information for the nodes because such edge information characterizes the natural interaction between the connected nodes~\cite{gilmer2017neural,gong2019exploiting,dwivedi2020generalization}. For example, an edge can record the bond types in a molecular graph, the rating scores in a user-movie social media graph, or the contextual information in our customer-skill graph in Alexa, etc. To improve the performance of our model, we explore two edge-based convolution operators.

\textbf{\emph{NNConv:}} NNConv is proposed in the Message Passing Neural Network designed to infer quantum properties of molecules based on the structure of the molecules and their bond types (e.g. edge features)~\cite{gilmer2017neural}.
It demonstrates the superior performance to the corresponding baselines in terms of the inferences on quantum properties by considering the edge features in GCN message passing stage.
Analogous to their setup, where customer and skill nodes refer to the vertices of molecules and utterances refer to the the bond types, we choose to apply this module in the proposed PDRFE model.
For completeness, the message passing function is shown in Eq.~\ref{NNConv}:
\begin{equation}
\label{NNConv}
\textbf{h}_{i}^{l+1} = \textbf{h}_{i}^{l} + \mathrm{mean}_{j\in \mathcal{N}(i)}\left(\left\{(W_e\cdot \textbf{e}_{i,j} + b_e) \textbf{h}_j^{l} \right\}\right),
\end{equation}
where $\mathrm{mean}(\cdot)$ refers to the mean aggregation function in GCN. 
$h_i^l$ refers to the embedding of $i$-th node at layer $l$. $W_e$ and $b_e$ are the linear transformation parameters for the input edge features $\textbf{e}_{i,j}$. $\mathcal{N}(i)$ refers to the neighboring node set to the node $i$.

\textbf{\emph{Edge-based attention:}}. In GCNs, it is well known that attention allows for the efficient assignments of different weights to the connected edges of a node, which endows the interpretability on the importance of those edges to that node~\cite{velivckovic2017graph,wang2019kgat,wang2019heterogeneous}.
In our case, since utterances characterize customer-skill interactions, and attention is used to determine the weights of such interactions, we can also introduce the utterance factor into the attention module to control the generation of attention weights. This way, the attention weights would reflect the importance of such utterance as compared to all other utterances connected to the same node. Please refer to Eq.~\ref{GATConv1}-\ref{GATConv3} below for details:

\begin{equation}
\label{GATConv1}
\textbf{h}_i^{l+1} = \sum_{j\in \mathcal{N}(i)} \alpha_{i,j} W^{l} \textbf{h}_j^{l},
\end{equation}

\begin{equation}
\label{GATConv2}
 \alpha_{i,j}^{l} = \mathrm{softmax} (k_{i,j}^{l}),
\end{equation}

\begin{equation}
\label{GATConv3}
 k_{i,j}^{l} = \mathrm{LeakyReLU}\left(\vec{a}^T [W_a[ \textbf{h}_{i}^l\|\textbf{e}_{i,j}] \| W_a[\textbf{h}_{j}^l\|\textbf{e}_{i,j}]]\right),
\end{equation}
where $W_a$ and $W^l$ are the linear encoding matrix for attention module and for message passing at layer $l$, respectively. $\alpha_{i,j}$ and $\vec{a}$ refer to the attention weights and linear encoding vector for attention module, respectively. $[\cdot \| \cdot]$ refers to the concatenation operator.

\subsection{Personalizer}
The typical output of GCN model is node embeddings.
However, these node embeddings are too generic to be used for context-sensitive applications as ours.
For example, the link prediction objective, shown in Section~\ref{problem_define}, predicts the edge between a customer and a skill using simple inner product without having any context considerations. To resolve this problem, we design a new mechanism to provide the context-tailored customer embedding based on the learned customer embedding and utterance.
We denote this mechanism as a \textit{Personalizer}.
Formally ~\sz{defined as follows},
\begin{equation}
\label{eqn:personalizer}
\textbf{h}_s^p = f_p([\textbf{h}_s^l\|utt])
\end{equation}
where $\textbf{h}_s^p$ is context-considered customer embedding, $f_p$ is a \textit{personalizer} function, and $utt$ is the utterance text that we would like to consider. 

In this paper, we modeled $f_p$ to be a 2-layer neural network. This personalized embedding $\textbf{h}_s^p$ replaced the conventional context-free customer embedding $\textbf{h}_u^L$ in link prediction objective (Section~\ref{problem_define}), as well as replacing the customer embedding in the downstream task. This \textit{Personalizer} also works with parallel edges where the customer node is connected to the skill node via multiple different utterances. 
Note that, this mechanism is trained during the representation learning time and not the downstream task training time.

\section{Experiment Setup \& Results}
\label{res}

In this section, we describe how we setup our experiment and the dataset for training/evaluation for both representation learning and downstream evaluation (Section~\ref{dataset}), ~\sz{define} the baseline models to compare our work with (Section~\ref{baseline}),  ~\sz{perform} hyper-parameter tuning (Section~\ref{hyperparameter_setup}). We then present the results of our best performing model, and provide ablation study on the modules proposed in the previous section to demonstrate the effects of each of the modules to the downstream task.

\subsection{Dataset Setup} \label{dataset}
In this section, we provide the details of the dataset preparation for both representation learning (a graph dataset) and downstream evaluation (a regular input-output dataset). Both datasets are generated from the raw customer-skill interaction logs that are randomly collected to remove biases during data collection. For privacy, we only provide a fake example for illustration purposes (see Table~\ref{fake_data_tab}). Additionally, we include metadata come with the interaction logs in Table~\ref{meta_data_tab}. These metadata will be used either in generating the initial node embeddings or for edge feature assignments. Note that the ``Defect" in Table~\ref{meta_data_tab} is an 0-1 binary label used for $y$, which is predicted by CPDR team, in the downstream evaluation.

\begin{table*}[ht] 
\hspace*{-4cm}
  \caption{Metadata in the Customer-Skill Interaction Logs}
  \label{meta_data_tab}
  \centering
\resizebox{\textwidth}{!}{
\begin{tabular}{lllll}
\toprule
Name of Metadata      & Description                                                                            & Cardinality & Example                                                                               & Usage                                                                                                  \\ \midrule
Skill\_category       & Category of skills                                                                     & 22          & \begin{tabular}[c]{@{}l@{}}Social / \\ Education \& reference\end{tabular}            & \multirow{4}{*}{\begin{tabular}[c]{@{}l@{}}For initializing\\ skill node\\ embeddings\end{tabular}}    \\
Skill\_type           & Skill type                                                                             & 7           & CUSTOM                                                                                &                                                                                                        \\
Skill\_subcategory    & Subcategory of skills                                                                  & 69          & \begin{tabular}[c]{@{}l@{}}Education \& Reference / \\ Social Networking\end{tabular} &                                                                                                        \\
Reporting\_category   & General category                                                                       & 28          & Education / Social                                                                    &                                                                                                        \\ \hline
wbr\_cor              & Weekly business review country                                                         & 19          & US / ROW                                                                              & \multirow{4}{*}{\begin{tabular}[c]{@{}l@{}}For initializing\\ customer node\\ embeddings\end{tabular}} \\
is\_prime             & Prime membership?                                                                      & 2           & Y / N                                                                                 &                                                                                                        \\
is\_amu               & Amazon music subscriber?                                                               & 2           & Y / N                                                                                 &                                                                                                        \\
is\_smart\_home\_cust & Smart home customer ?                                                                  & 2           & Y / N                                                                                 &                                                                                                        \\ \hline
Utterance             & Utterances (users’ requests)                                                           & N/A         & Play tiktok                                                                           & \multirow{2}{*}{Edge features}                                                                         \\
Defect                & \begin{tabular}[c]{@{}l@{}}Whether a request is\\  fulfilled successfully\end{tabular} & 2           & 1 / 0                                                                                 &                                                  \\ \bottomrule                                                     
\end{tabular}}
\end{table*}

\subsection{Baselines}
\label{baseline}
In order to verify the performance of our proposed model, we compare it with 4 baseline models. Here we list all these models in Table~\ref{baseline_tab} with reasons of selection.

\begin{table*}[ht]
  \caption{Baseline Models}
  \label{baseline_tab}
 \resizebox{\textwidth}{!}{
\begin{tabular}{l|ll}
\toprule
Model Name                                                                                & Description                                                                                                                                                                    & Reason of the Choice                                                                                                      \\ \hline
PinSage~\cite{ying2018graph}                                                                                   & \begin{tabular}[c]{@{}l@{}}PinSage is a large scale GCN model designed\\  by Pinterest for recommending pins (images) \\ to people. Considered only one node type\end{tabular} & \begin{tabular}[c]{@{}l@{}}Representation learning model\\  that has been applied to \\ real-world scenarios\end{tabular} \\ \hline
Revised PinSage                                                                           & \begin{tabular}[c]{@{}l@{}}Extend PinSage to 2 types of nodes\\  (both customer and skill)\end{tabular}                                                                        & \begin{tabular}[c]{@{}l@{}}Improve the original PinSage\\  for comparison in our case\end{tabular}                        \\ \hline
\begin{tabular}[c]{@{}l@{}}Rational Graph \\ Convolutional Network \\ (RGCN)~\cite{schlichtkrull2018modeling} \end{tabular} & Simple GCN designed for link prediction                                                                                                                                        & Baseline for link prediction                                                                                              \\ \hline
Raw One-hot Encoding                                                                      & \begin{tabular}[c]{@{}l@{}}No Training, just use one-hot encoder on\\  the user/skill metadata and use these for\\  the downstream task\end{tabular}                           & \begin{tabular}[c]{@{}l@{}}Baseline for non-graph-based \\ approach\end{tabular}                                                                                      \\ 
\bottomrule
\end{tabular}}
\end{table*}

\subsubsection{Representation Learning}
\label{RL}

To prepare the graph for representation learning, firstly we subsample our dataset following these three steps to obtain a subsampled interaction log: 1) Randomly subsample 10\% customers by their customer ID (CID). 2) Collect all the interactions corresponding to the subsampled CIDs in 14-day logs. Then we turn this log into a graph by treating each interaction log as an edge in the graph connecting the corresponding customer and skill nodes. Overall, the processed graph contains 107,480,710 number of customer-skill interactions (e.g. edges) for 12,031,327 customers (e.g. customer node) and 8,110 skills (e.g. skill node). 

To train and evaluate the representation model, we randomly split this graph to two subgraphs based on uniform edge sampling. We sample 80\% edges for training graph and the other 20\% for testing graph.
\begin{table}[]
\centering
\caption{Legend for Reference Models.}
 \label{ben_model_tab}
\begin{tabular}{@{}ll@{}}
\toprule
\multirow{2}{*}{\begin{tabular}[c]{@{}l@{}}Reference Model\\ Configuration\end{tabular}} & \multirow{2}{*}{Description}                             \\
                                                                                         &                                                          \\ \midrule
RGCN                                                                                     & The backbone RGCN model                                  \\
EC                                                                                       & The backbone RGCN + NNConv                               \\
EAtt                                                                                     & The backbone RGCN + edge-based attention                 \\
Per                                                                                      & The backbone RGCN + \textit{Personalizer} \\ \bottomrule
\end{tabular}
\end{table}

\begin{table*}[ht]
  \caption{Comparison between Our Best Models with the Baselines. Evaluation is based on the testing cross-entropy in the downstream evaluation. \textbf{Bold} numbers refer to the best results.}
  \label{overall_tab}
  \centering
\begin{tabular}{lcc}
\toprule
\multirow{2}{*}{Models}      & \multicolumn{2}{c}{Cross-Entropy (lower the better)} \\ \cline{2-3} 
                                & \multicolumn{1}{l}{Logistic Regression} & \multicolumn{1}{l}{2-layer Neural Network}     \\ \midrule
PinSage (Baseline)              & 0.543                   & 0.441                      \\
Revised PinSage (Baseline)      & 0.519                   & 0.421                      \\
Raw One-Hot Encoding (Baseline) & 0.462                   & 0.424                      \\
RGCN (Baseline)                 & 0.372                   & 0.371                      \\
PDRFE (w/ ``NNConv'')            & \textbf{0.317}                   & \textbf{0.303}                   \\
PDRFE (w/ edge-based attention)  & \textbf{0.324}              & \textbf{0.302}  \\
\bottomrule
\end{tabular}
\end{table*}

To conduct the experiments, we use an Amazon AWS cluster with V100 NVIDIA GPUs with the following libraries: PyTorch~\cite{NEURIPS2019_bdbca288}, DGL~\cite{wang2019deep} and ``"torch-two-sample"~\cite{torch_two_sample}.

\subsubsection{Downstream Evaluation}
Once the representation model is trained, we would generate embeddings for all the nodes, including personalized embeddings. We then convert the graph back into the customer-skill interaction log table with the only change udated to the personalized embeddings for given customer, skill and the corresponding utterance (e.g. $\textbf{e}_{i,j}$). We randomly split the dataset into train/validation/test sets in the standard way with split ratios being 6:2:2. The data are then evaluated with the proposed downstream objective (see Section~\ref{problem_define}).

\subsection{Hyperparameter Setup}
\label{hyperparameter_setup}
The following hyperparameter setup (shown in Table~\ref{hyperparameter_tab}) is applied for the representation learning and the downstream evaluation. 
For the downstream evaluation, we use two classifiers: 1) logistic regression module being the linear classifier 2) A nonlinear classifier which consists of a 2-layer neural network with ReLU activation function. The hidden dimension for this network is 32. We choose to use a small number for the hidden dimension to avoid overfitting of this nonlinear classifier, ~\sz{which can improve the quality of learned representation.}

\begin{table}[ht]
  \caption{Hyperparameter Steup}
  \label{hyperparameter_tab}
  \centering
\begin{tabular}{lcc}
\toprule
\multirow{2}{*}{Hyperparameter}                             & \multicolumn{2}{c}{Model}                                                                                                                                                \\ \cline{2-3} 
                                                            & \multicolumn{1}{l}{\begin{tabular}[c]{@{}l@{}}Representation \\ Learning\end{tabular}} & \multicolumn{1}{l}{\begin{tabular}[c]{@{}l@{}}Defect\\ Prediction\end{tabular}} \\ \midrule
Batch Size                                               & \begin{tabular}[c]{@{}c@{}}512/1024/\\ 2048/4096 \\ ($\times$5 for negative sampling)    \end{tabular}                          & 256                                                                             \\
Learning Rate                                               & 10\textasciicircum{}\{-4\}                                                             & 10\textasciicircum{}\{-4\}                                                      \\
\begin{tabular}[c]{@{}l@{}}Hidden \\ Dimension\end{tabular} & 128                                                                                    & 32                                                                              \\
Total \# of Epochs                                          & $\leq$ 10                                                                              & \begin{tabular}[c]{@{}c@{}}When validation \\ loss is stable\end{tabular}      \\ \bottomrule
\end{tabular}
\end{table}

\begin{figure*}[ht]
     \centering
     \begin{subfigure}[b]{0.45\textwidth}
         \centering
         \includegraphics[width=\textwidth]{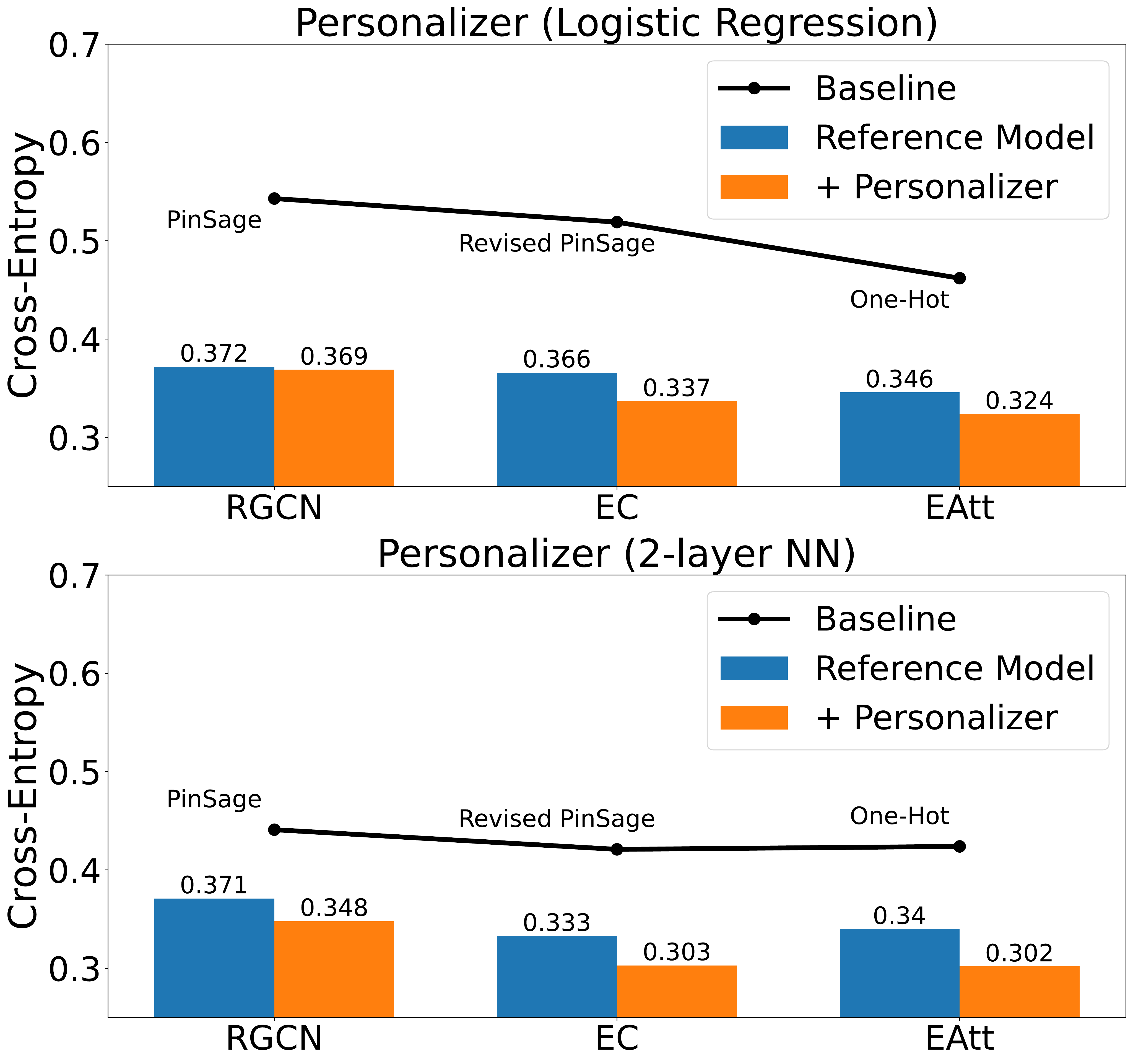}
         \caption{\textit{Personalizer}}
         \label{ablation:personalizer}
     \end{subfigure}
     \hfill
     \begin{subfigure}[b]{0.45\textwidth}
         \centering
         \includegraphics[width=\textwidth]{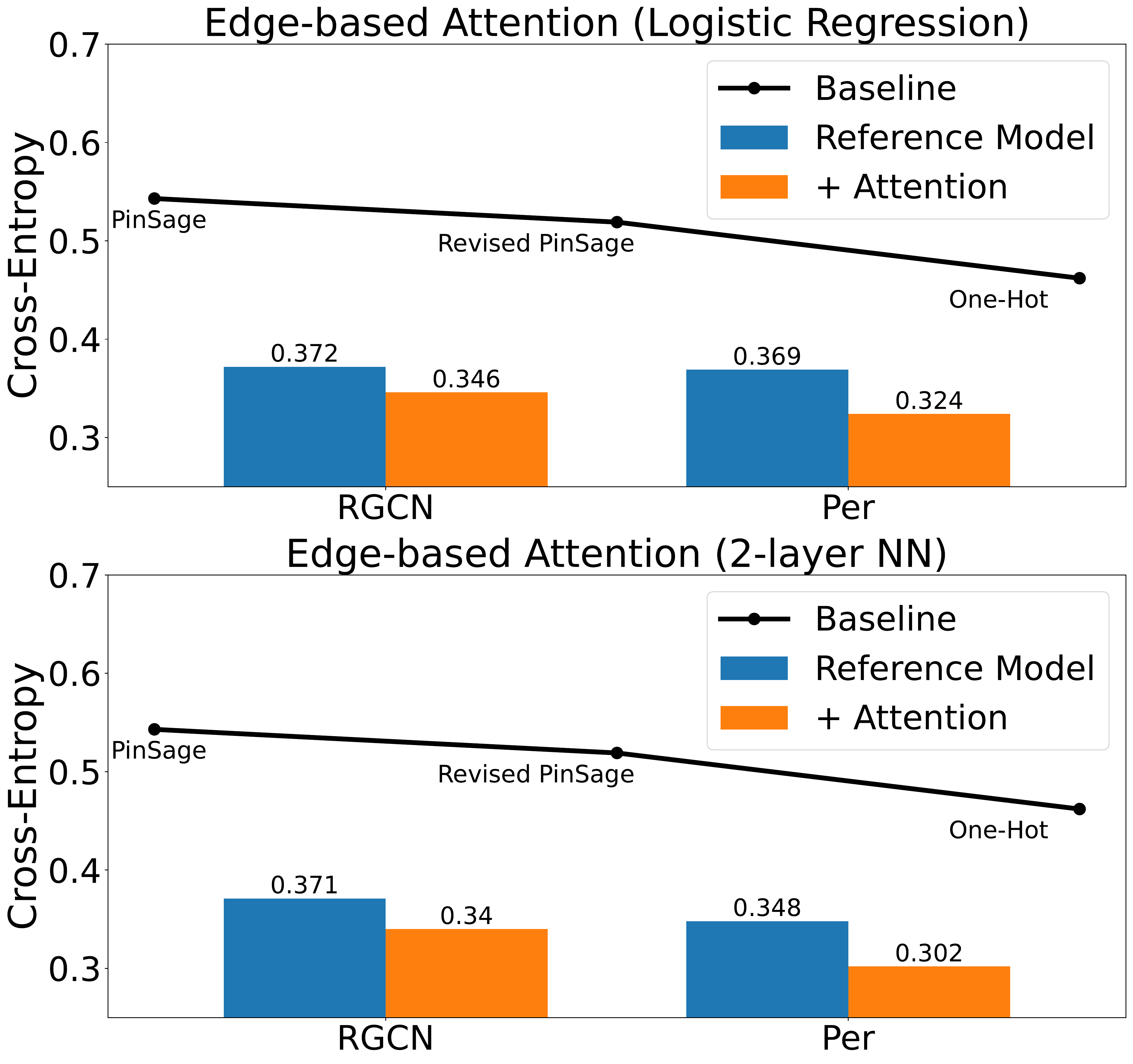}
         \caption{Edge-based attention}
         \label{ablation:attention}
     \end{subfigure}
     \hfill
     \begin{subfigure}[b]{0.45\textwidth}
         \centering
         \includegraphics[width=\textwidth]{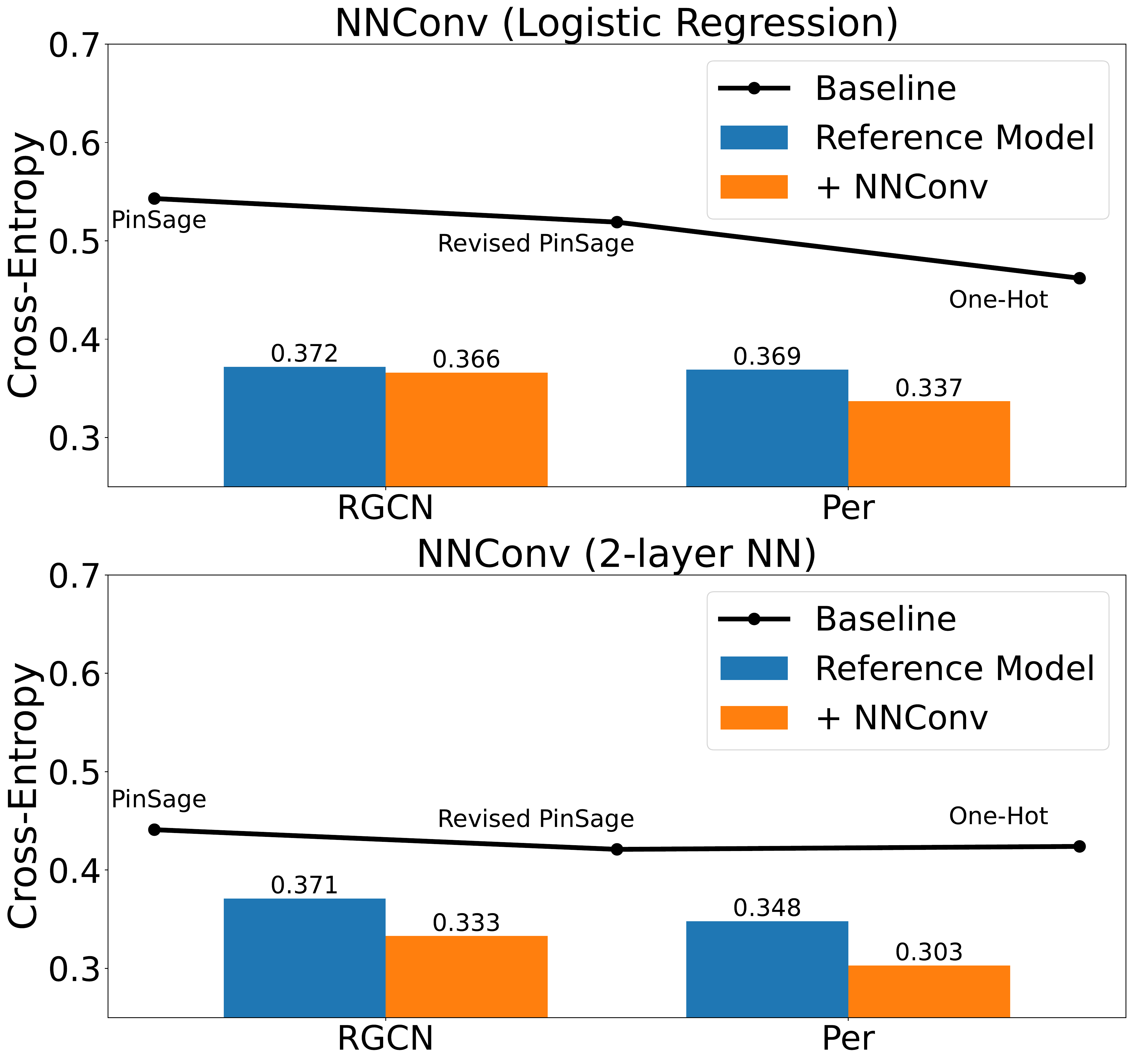}
         \caption{``NNConv''}

         \label{ablation:edge_conv}
     \end{subfigure}
     \hfill
        \caption{Ablation study on modules in PDRFE. The 3 bar-plots include the comparisons between the models with and without the modules discussed in the method section. Specifically, there are 3 modules included in this comparison: 1. The \textit{personalizer}; 2. The NNConv; 3. The edge-based attention. The reference models (in {\color{blue} blue}) are the models without the component, which ~\sz{are} compared to the models with the component (in {\color{orange} orange}). The numbers reported in the figures are cross-entropies. The sub-captions indicate which module in PDRFE we compare with, and the tick labels indicate the name of the reference models with details in the legend Table~\ref{ben_model_tab}.}
        \label{ablation_study_fig}
\end{figure*}

\subsection{Model Comparisons}
\label{model_comparisons}

Here we present the comparison between the PDRFE model and the 4 baseline models. Results are evaluated according to the testing cross entropy for the downstream defect prediction task (refer to Table~\ref{overall_tab}). It is clear that the PDRFE models (with either the ``NNConv'' or the edge-based attention) obtain lower cross-entropy. The relative improvements in cross-entropy ranges between 15\% to 41\% for logistic regression model, and 19\% to 31\% for the 2-layer neural network model. In addition, it is obvious that the incorporation of the utterance embeddings into the model contribute to huge improvements for the downstream task. On one hand, the edge features improve the precision for node representations. On the other hand, the \textit{personalizer} resolves the two aforementioned ambiguities. Both components contribute to the out-performance of PDRFE over the baseline models.

\subsection{Ablation Study}
\label{ablation_study}
To study the individual effects of each module in PDRFE, we perform this ablation study on the 4 modules in PDRFE. Results are shown in Figure~\ref{ablation_study_fig}. It is shown that in all cases in this ablation study, the models that have the proposed components in PDRFE outperform the reference models where such components are removed. Specifically, the maximum performance boost for the 3 components are: 1) \textit{Personalizer}: 12\% 2) ``NNConv'': 19\% 3) Edge-based attention: 13\%. Each component can contribute to improvements in the defect prediction. However, when consider them jointly in PDRFE model, the performance is even better. This indicates the necessity of incorporating the edge features into the message passing as well as the \textit{personalizer} in the GCN model.

\section{Conclusion}
\label{conclusion}
In this work, we build a heterogeneous graph with attributed edges based on customers' past interactions with the invoked skills. Besides, we propose PDRFE, a GCN-based representation learning model, designed for generating personalized customer-skill embeddings from the built graph. The proposed model is evalulated according to a binary classification downstream task, known as defect prediction, which reflects the customer's preferences to the skills. We compare our models to the proposed baselines and observed up to 41\% improvement in the cross-entropy in the downstream evaluation. Additionally, an ablation study is included to understand the performance contribution of the modules in PDRFE where the resulting improvements range from 12\% to 19\%.

Our PDRFE model impacts customer experience and improves downstream customer satisfaction by reducing the defect rate during fulfillment of a user's request by Alexa's skills. We demonstrate that PDRFE model generates more precise customer and skill representations which ~\sz{encode} customer-skill relationship, resulting in lower cross-entropy in the defect prediction. As a result, such characterization of customer-skill connectivity by PDRFE can be possibly applied to the existing Alexa DR tools in order to improve the accuracy of the current routing services. For example, it can benefit Alexa DR's ranking model (HypRank~\cite{kim2018scalable}) which considers joining the information from the user side such as utterance, interpretation, etc. and forms hypothesis to re-rank the existing skill candidates to find the most likely skill to answer ~\sz{a} user’s request. It reduces the defect rate while answering users’ requests. However, this model does not consider the customer and skill interactions as a graph, which means the current representations in HypRank may not reflect the natural interactions between customers and skills. In the future work, we plan to inject the representations from the proposed PDRFE model to provide more contextual information for the customers and skills to reflect their natural connectivity into HypRank ranking decisions.

\small
\bibliographystyle{ACM-Reference-Format}
\bibliography{reference.bib}

\end{document}